\documentclass[runningheads]{llncs}

\usepackage[T1]{fontenc}

\usepackage[hidelinks]{hyperref}
\usepackage{xcolor}

\usepackage{amsmath}
\usepackage{xfrac}
\usepackage{gensymb}
\usepackage{fixmath}

\usepackage{array}

\usepackage{comment}
\usepackage{caption}
\usepackage{subcaption}

\usepackage{graphicx} 

\title{Position and Altitude of the Nao Camera Head from Two Points on the Soccer Field \\ plus the Gravitational Direction}
\titlerunning{Position and Altitude of the Nao Robot from Two Points}
\author{Stijn Oomes\inst{1} \and Arnoud Visser\inst{2}\orcidID{\href{https://orcid.org/0000-0002-7525-7017}{\includegraphics[scale=0.05]{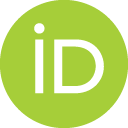}}}
}
\authorrunning{S. Oomes \and A. Visser}
\institute{
Oomes Vision Systems, Amsterdam, NL
\and
Intelligent Robotics Lab, Universiteit van Amsterdam, NL
\hspace{5em}
\href{https://www.intelligentroboticslab.nl/}{www.intelligentroboticslab.nl}}

\date{February 2024}

\begin{document}

\maketitle

\begin{abstract}
To be able to play soccer, a robot needs a good estimate of its current position on the field. Ideally, multiple features are visible that have known locations. By applying trigonometry we can estimate the viewpoint from where this observation was actually made. Given that the Nao robots of the Standard Platform League have quite a limited field of view, a given camera frame typically only allows for one or two points to be recognized.

In this paper we propose a method for determining the $(x, y)$ coordinates on the field and the height $h$ of the camera from the geometry of a simplified tetrahedron. This configuration is formed by two observed points on the ground plane plus the gravitational direction. When the distance between the two points is known, and the directions to the points plus the gravitational direction are measured, all dimensions of the tetrahedron can be determined.

By performing these calculations with rational trigonometry instead of classical trigonometry, the computations turn out to be 28.7\% faster, with equal numerical accuracy.

The position of the head of the Nao can also be externally measured with 
the OptiTrack system. The difference between externally measured and internally predicted position from sensor data gives us 
mean absolute errors 
in the 3-6 centimeters range, when we estimated the gravitational direction from the vanishing point of the outer edges of the goal posts.

\end{abstract}

\section{Introduction}

 RoboCup is an event where robot capabilities are tested by playing a game of soccer between one team against another \cite{kitano1997robocup}. Already from the beginning of the competitions, the importance of visual localization was recognized \cite{Iochhi2000,Enderle2001}. Since that start methods like Visual Odometry have been developed, but those methods rely on enough dense and distinctive textures \cite{Alkendi2021}. In the RoboCup often not much more than the green turf is visible, with an occasional white line (if not obstructed by other robots nearby). When a L-, T- or X-intersection  on a white line-crossing can be observed \cite{Laue2009}, a distinctive point on the field has been recognized (although due to the symmetry of the field unfortunately not unique). Two of those distinctive points are enough to estimate  the location where this observation is made, as demonstrated in this paper.

\section{ Rational Trigonometry 
}

To understand the relations between observable variables (measurements with one or more cameras) and not directly observable variables (3D properties of objects and their configurations) transformations have to be defined. Rational Trigonometry \cite{wildberger2005divine} uses quadrances and spreads instead of distances and angles. The quadrance is defined as the quadratic sum of differences in the coordinate values of two points. The spread is defined as a proportion of quadrances; for a right triangle it is the opposite quadrance divided by the hypotenuse quadrance.  
Both quantities can be easily converted back and forth to the well-known concepts of distance and angle, although they are defined \textit{independently}. A \textbf{quadrance} $Q$ is the square of the distance $d$. This makes it in essence a measure of area in Euclidean space. Quadrances have dimension $m^2$.

\begin{equation}
    Q = d^2
\end{equation}

A spread measures the divergence between two lines. A \textbf{spread} $s$ is the square of the sine of an angle $\alpha$. 

\begin{equation}
    s = \sin^2 \alpha
\end{equation}

Spreads  have a range between $[0 - 1]$ and are dimensionless, but is actually the ratio of quadrances (so dimension $\sfrac{m^2}{m^2}$). This characteristic has important consequences, because this range is better balanced than the ranges of degrees ($[0 - 360\degree]$) and radians ($[0 - 2\pi]$) once translations and rotations have to be combined. It also means that the spread has no sign, and is actually the same value for adjacent angles. This prevents many ambiguity checks in the calculations.

\vspace*{-1em}
\begin{figure}[!htb]{}
     \centering
     \includegraphics[width=0.5\textwidth]{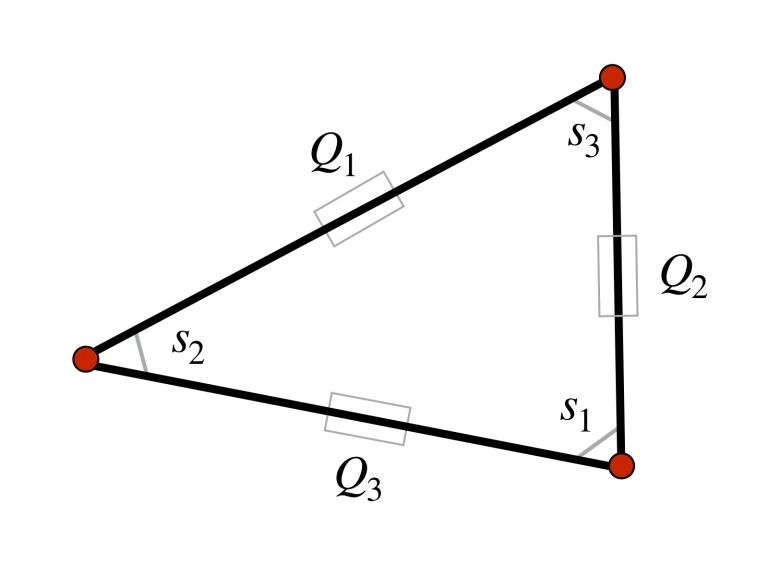}
     \caption{The quadrances $Q_i$ and spreads $s_i$ of a triangle.}
     \label{fig:triangle}
\end{figure}

The relation between quadrances and spreads of triangle, as illustrated in Fig.~\ref{fig:triangle} is given by the \textbf{Spread law} of rational trigonometry:
\vspace*{-0.5em}
\begin{equation}
    \frac{s_1}{Q_1} = \frac{s_2}{Q_2} = \frac{s_2}{Q_2}
    \label{eq:spread_law}
\end{equation}

Yet, for the method applied in this paper mostly the \textbf{Cross law} of rational trigonometry is used:
\vspace*{-0.5em}
\begin{equation}
    (Q_1 + Q_2 - Q_3)^2 = 4 Q_1 Q_2 (1-s_3)
    \label{eq:cross_law}
\end{equation}

The Cross law (Equation~\ref{eq:cross_law}) can be used to calculate an unknown quadrance $Q_3$ from two known quadrances $Q_1$ and $Q_2$ and the spread $s_3$ between those two quadrances (and opposite the unknown quadrance $Q_3$). Once $s_3$ and all three quadrances $Q_{1\cdots3}$ of a triangle are known, the other two spreads $s_1$ and $s_2$ can be calculated from the Spread law (Equation~\ref{eq:spread_law})

\section{Method}

The first phase is the detection of the relevant objects in the world, that is, on the soccer field. Those are the white lines on the green field, especially the X-, T-and L-intersections. Of course there are also objects like the goals, the ball and the other robots; team-members and opponents are distinguishable by their different colored jerseys \cite{Owusu2024thesis}.
The distance between two of those points detected on the field can be used to estimate the position where those points are observed from the camera position. The position of the X-, T-and L-intersections are specified in the Standard Platform League rule book \cite{spl2024rules}, so their distances are known.

For efficiency reasons in RoboCup typical scan-line techniques are used to detect lines, searching for regions of white pixels surrounded with green pixels \cite{Laue2009}.  Yet, many other white objects are present on the field (such as goal-posts, the ball and other robots), so additional filters are needed to reduce the number of false positives \cite{Farazi2015monocular}. Instead of hand-crafting such filters for the lines and their crossings, we used a state-of-the-art detector. 

\begin{figure}[!htb]{}
     \centering
     \includegraphics[width=0.5\textwidth]{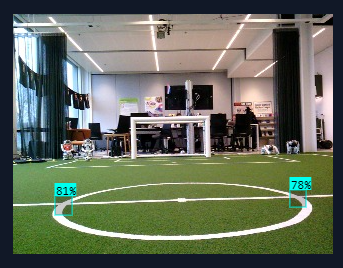}
     \caption{The X-intersection detected on the Standard Platform League field.}
     \label{fig:intersections}
\end{figure}

The X-, T-and L-intersections in this paper are detected with the detector based on YOLO v8, a detector which is slightly easier to fine-tune on new types of objects than YOLO v9 \cite{wang2024yolov9}. The detector is based on the YOLO v8 XL model pretrained on the COCO dataset, further finetuned on 1309 L-intersections, 1229 T-intersections and 806 X-intersections. This intersections were already annotated in the TORSO-21 dataset \cite{TORSO-21}, but further extended with annotated mages recorded by the Dutch Nao Team \cite{DNT2020synthetic}. 
This dataset, including the YOLO v8 XL model trained on this data, is publicly available\footnote{ \url{https://universe.roboflow.com/fieldmarks/splfieldmark}, courtesy Gijs de Jong}. This model achieved a precision of 86.1\% and a recall of 84.2\%. An example of such detection of two X-intersections can be seen in Fig.~\ref{fig:intersections}. The intersections further away are also detected, but with a too low confidence to be displayed (a confidence below 50\%). From the rule book it is known that the distance between the two X-intersections is 1.5m.

\vspace*{-0.5em}
\subsection{Position from 3 known points}

When three points are known in the world, one can formulate a third degree polynomial which once solved gives 4 solutions. This is known as the \textbf{P3P} problem, which was first solved by Grunert \cite{grunert1841pothenotische} and further refined by Finsterwalder \& Scheufele \cite{finsterwalder1903ruckwartseinschneiden}. Grunert solution is nicely described by \cite{haralick1991analysis}, which starts with the following three equations generated by the law of cosines:
\begin{equation}
\begin{split}
x^2 + y^2 - 2 x y \cos(\alpha) = a^2 \\
x^2 + z^2 - 2 x z \cos(\beta) = b^2 \\
y^2 + z^2 - 2 y z \cos(\gamma) = c^2
\end{split}
\end{equation}
Finsterwalder has shown that this is equivalent with solving $\lambda$ in this third degree polynomial for the P3P problem:
\begin{equation}
\begin{split}
&\lambda^3 \phantom{i} c^2 (c^2 \sin^2 \alpha - a^2 \sin^2 \gamma ) \phantom{i} + \\
&\lambda^2  \phantom{i}
[a^2 (b^2-a^2) \sin^2 \gamma 
- c^2 (2 b^2+ c^2) \sin^2 \alpha 
+ 2 a^2 c^2(1- \cos \alpha \cos \beta \cos \gamma )
] \phantom{i}+  \\
&\lambda \phantom{i}
[a^2 (a^2-c^2) \sin^2 \beta 
+ b^2 (b^2+ 2 c^2) \sin^2 \alpha
- 2 a^2 b^2(1- \cos \alpha \cos \beta \cos \gamma )
] \phantom{i}+  \\
&b^2 [a^2 \sin^2 \beta  - b^2 \sin^2 \alpha ]
 = 0
\end{split}
\end{equation}

After the overview of Haralick \cite{haralick1991analysis} several other researchers came with solutions for the P3P problems, such as \cite{gao2003complete} and \cite{Kneip2011cpvr}. 

When a fourth point is known one can disambiguate between the four solutions, which makes P3P a special case of the PnP problem with $n$ the number of known points. Yet, in our case only two points are known: $n=2$.


\subsection{Position from 2 points on the field and gravity}

The two point problem can be solved if we have a robot 
and a measure of the direction of gravity. 
Actually, there are two ways to get a measure on the direction of gravity. Initially we used the measurement of the an Inertial Measurement Unit (IMU) on-board of the robot. By also using the measurement values of the pitch and yaw angles of the head relative to the torso, this direction is transformed to the coordinate frame of the camera. Yet, together with timing errors small directional deviations resulted in positional errors in the range of 8-25 centimeters range. As a check, we also derived the gravitational direction from the perspective in the image, from the vanishing point of the outer edges of the goal posts. Here we found that the mean absolute errors are much better in the 3-6 centimeters range. So, for the rest of this paper we use the gravitational direction from the perspective, although this means that this direction can only be estimated when parallel verticals structures are visible.  With better calibration of the gravitational direction from the IMU and the kinetic chain, we could skip this intermediate step.

\begin{figure}
    \centering
    \includegraphics[width=0.925\linewidth]{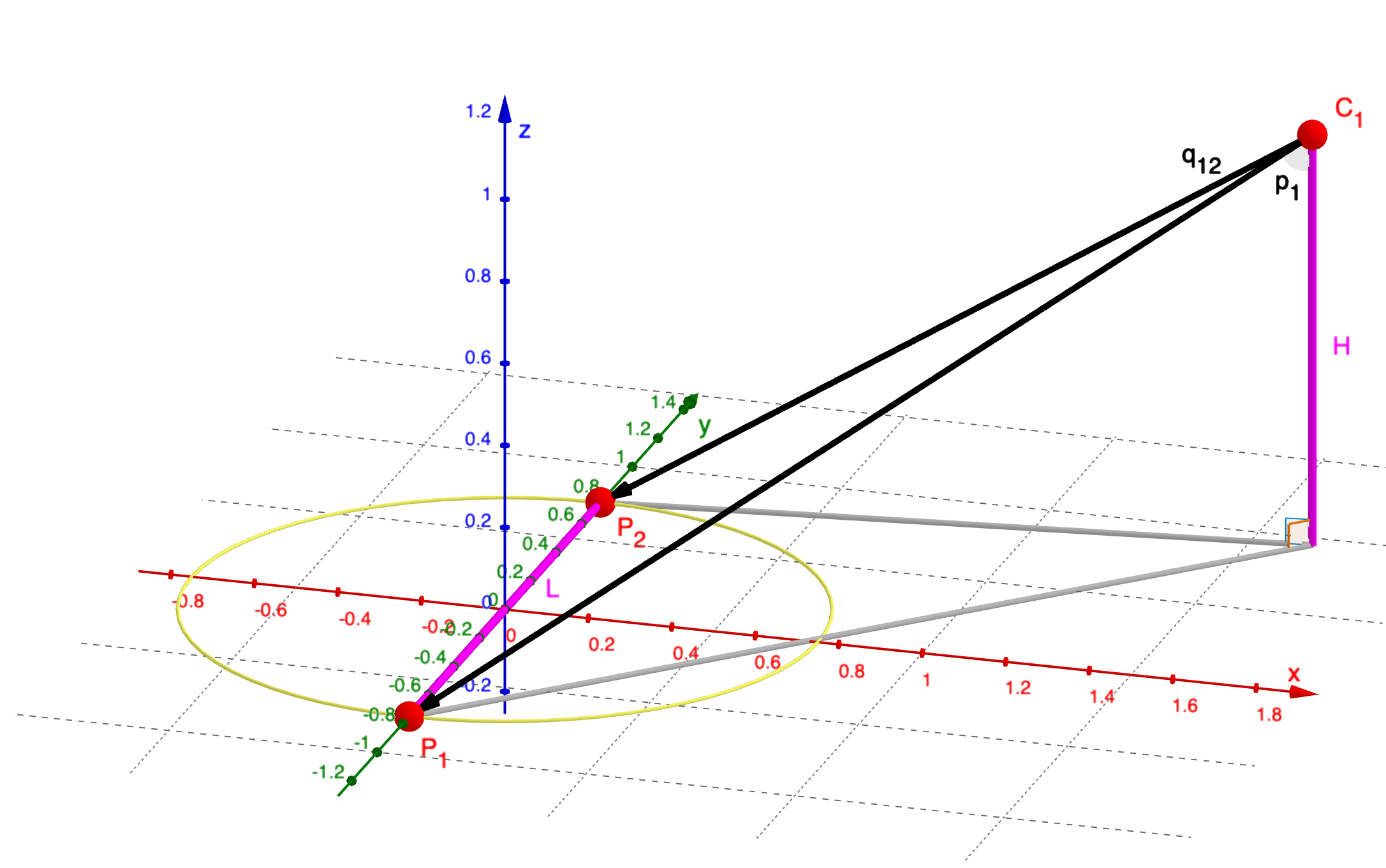}
    \caption{A line $L$ between two known points seen from a camera. The red points on the ground plane are the intersections of middle line and circle of which the quadrance $L$ is known. The red point on the top right is the camera point. Together with the foot of the altitude these points form a tetrahedron with two right angles. The altitude quadrance $H$ can be determined from quadrance $L$ and the spreads $p_1$, $p_2$, and $q_{12}$.}
    \label{fig:tetrahedron}
\end{figure}

For two points on the soccer field we know the distance, so also the quadrance $L$. We can determine their directions in camera coordinates from the projection in the image. With the direction of gravity $g$, we can convert those two directions to the spreads $p_1$ and $p_2$ between the visual directions and the gravitational direction $g$. The three vectors form a tetrahedron with the camera on top, as illustrated in Fig. \ref{fig:tetrahedron}. The spread between the two visual directions is $q_{12}$. The quadrance $H$ is along the gravitational direction $g$ and that is the parameter to be determined.

\vspace*{-1em}

\vspace*{-0.5em}
The tetrahedron has 4 vertices, 6 edges, and 4 faces. So, in the tetrahedron 6 distances / quadrances can be distinguished (see the three colored pairs in Fig.~\ref{fig:tetrahedron}) and 12 angles / spreads. One quadrance $L$ and five spreads are known: $(p_1,p_2,q_{12})$ and the two right angles formed by $H$ and the ground plane. The other spreads and quadrances can be derived from those measurements.


From the three measured spreads $p_1$, $p_2$, and $q_{12}$  and the known quadrance $L$ we can determine the vertical height, denoted as the quadrance $H$:

\begin{equation}
    H = \frac
{(1-p_1)(1-p_2)}
{(1-p_1)+(1-p_2)-2 \sqrt{(1-p_1)(1-p_2)(1-q_{12})}}
L
\end{equation}

If we define P1 in Fig.~\ref{fig:tetrahedron} as the origin, and the Y-axis is the vector along line $L$ in the direction of P2,  the $(X,Y)$ coordinate where quadrance $H$ touches the ground plane can be calculated:

\begin{equation}
   X = 
\frac
{(1-p_1)(1-p_2)[q_{12}-(1-p_1)-(1-p_2)
+ 2 \sqrt{(1-p_1)(1-p_2)(1-q_{12})}]}
{[(1-p_1)+(1-p_2)
- 2 \sqrt{(1-p_1)(1-p_2)(1-q_{12})}]^2}
L
\end{equation}

\begin{equation}
 Y = 
\frac
{[(1-p_2)- \sqrt{(1-p_1)(1-p_2)(1-q_{12})}]^2}
{[(1-p_1)+(1-p_2)
- 2 \sqrt{(1-p_1)(1-p_2)(1-q_{12})}]^2}
L
\end{equation}

Note that in both equations no transcendental trigonometric functions like $\sin(\phi)$, $\arccos(\phi)$ and $tan(\phi)$ are used.

\subsection{Classical trigonometry}

The same method can also be formulated with distances and angles, which gives the following equations:

\begin{equation}
x = 
\frac
{\cos(\alpha_1) \cos(\alpha_2)
\sqrt{\sin^2(\beta_{12})-\cos^2 (\alpha_1)-\cos^2 (\alpha_2)+2 \cos (\alpha_1)\cos (\alpha_2)\cos (\beta_{12})}}
{\cos^2 (\alpha_1)+\cos^2 (\alpha_2)
- 2 \cos(\alpha_1)\cos(\alpha_2)\cos(\beta_{12})}
\phantom{i} l
\end{equation}

\begin{equation}
    y = 
\frac
{\cos^2(\alpha_2) - \cos(\alpha_1)\cos (\alpha_2)\cos (\beta_{12})]}
{\cos^2 (\alpha_1)+\cos^2 (\alpha_2)
- 2 \cos(\alpha_1)\cos(\alpha_2)\cos(\beta_{12})}
l
\end{equation}

\begin{equation}
    h = \frac
{\cos(\alpha_1)\cos(\alpha_2)}
{\sqrt{\cos^2(\alpha_1)+\cos^2(\alpha_1)-
2 \cos(\alpha_1)\cos(\alpha_2)\cos(\beta_{12})}}
l
\end{equation}

When optimized (removing redundant calculations), this algorithm requires 17x a transcendental function-call to calculate $\cos(\phi)$ and $\sin(\phi)$.


\section{Computational experiments}

Wildberger claims that rational trigonometry increases both the computational accuracy and the computational speed when compared to classical trigonometry \cite{wildberger2005divine}. Let us check whether we have found any evidence for this claim.

If we look at the accuracy of both computations we do not find any significant differences. The deviations start showing up around the 12th decimal. That is way beyond the measurement accuracy and precision for our position measurements. 

The code to calculate the camera positions from two points on the field with \textit{quadrances} and \textit{spreads} was executed multiple times on a Nao v6 robot, which indicates that one calculation takes on average $2.47 \cdot 10^{-05}$ sec. The implementation was in Python 2.7, the language supported by the standard Aldebaran NaoQi image, so could be further optimized in C++ or RUST. Yet, the Python implementation can already be used for a comparison with the same calculation with traditional method based on \textit{angle} and \textit{distance}, which needed for one calculation on average was  $3.18 \cdot 10^{-05}$ sec. This indicates that the rational algorithm on the Nao v6 robot is  $\approx 28.7\%$ faster, because it performs the calculations with only divisions and square-roots as computational heavy operations, in contrast with the trigonometric functions like $\cos(\phi)$ and $\sin(\phi)$ used in the classical method. 

\vspace*{-2em}
\begin{table}
\caption{Execution time of algorithm in seconds}
\begin{tabular}{ m{3.5cm} m{1.6cm} m{1.6cm} m{1.6cm} m{1.6cm} m{1.6cm} }
\textbf{\# runs} 
& 50x & 500x & 5000x & 50.000x & 500.000x \\
\textbf{classical algorithm} 
& $1.58\cdot10^{-3}$  & $1.59\cdot 10^{-2}$  & $1.61\cdot 10^{-1}$  & $1.59\cdot 10^{0}$  & $1.59 \cdot 10^{+1}$ \\
\textbf{rational algorithm} 
& $1.24\cdot10^{-3}$  & $1.24\cdot 10^{-2}$  & $1.23\cdot 10^{-1}$  & $1.23\cdot 10^{0}$  & $1.23 \cdot 10^{+1}$
\end{tabular}
\end{table}
\vspace*{-2em}

\section{Metrical experiments}

To see the relative location $(x,y,h)$ of the camera in the head of the Nao v6 robot relative to two points on the soccer field, we recreated two Simultaneous Localization and Mapping experiments, which were used as Technical Challenge in 2004 and 2005 \cite{spl2004challenge,spl2005challenge}.

The 2004 SLAM challenge specified five points: $[(160,100), (180, -30), (50,$
$-100), (-210,0), (-100,50)]$ - all cm from the origin at the center of the field. Those points were marked with red tape on the field. The 2005 SLAM challenge specified another five points: $[(130,120), (220, -150), 
(-160,-120), (-210,90),$
$ (270,0)]$, they were marked with black tape on the field. 
 
\begin{figure}[!htb]{}
    \centering
    \begin{subfigure}[t]{0.18\textwidth}
    \centering
    \includegraphics[height=0.085\textheight]{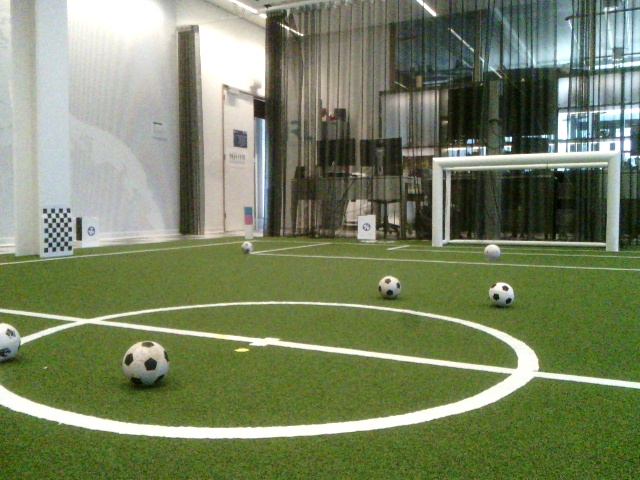}
    \caption{ \label{fig:red1}}
    \end{subfigure}
    \vspace*{1em}
    \begin{subfigure}[t]{0.18\textwidth}
    \centering
    \includegraphics[height=0.085\textheight]{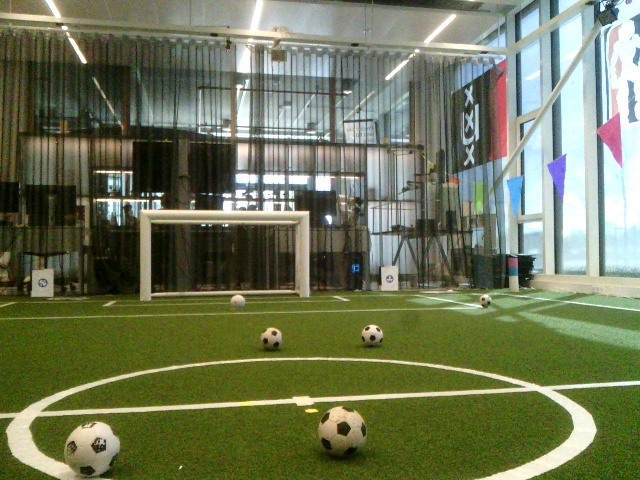}
    \caption{ \label{fig:red2}}
    \end{subfigure}
     \begin{subfigure}[t]{0.18\textwidth}
     \centering
    \includegraphics[height=0.085\textheight]{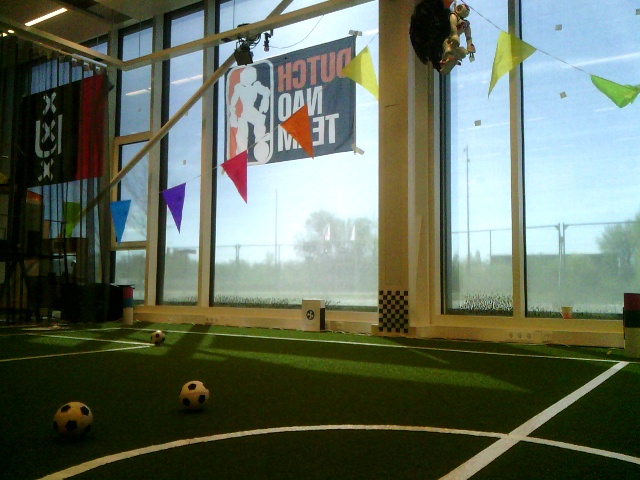}
    \caption{ \label{fig:red3}}
    \end{subfigure}
    \begin{subfigure}[t]{0.18\textwidth}
     \centering
    \includegraphics[height=0.085\textheight]{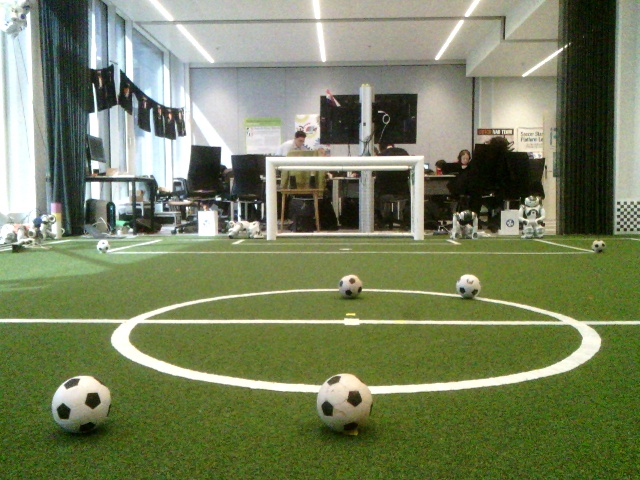}
    \caption{ \label{fig:red4}}
    \end{subfigure}
    \begin{subfigure}[t]{0.18\textwidth}
     \centering
    \includegraphics[height=0.085\textheight]{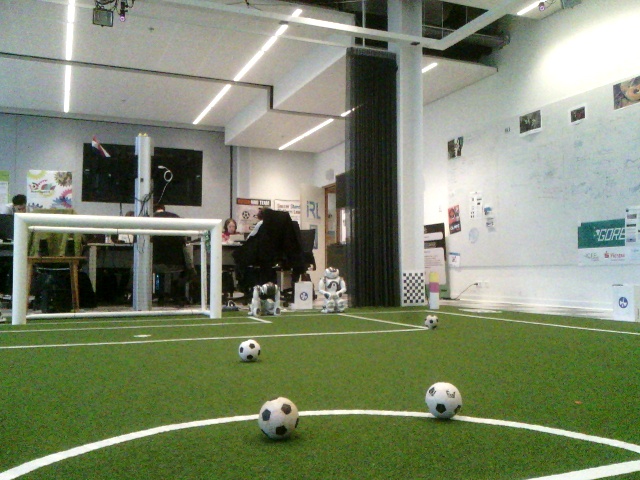}
    \caption{ \label{fig:red5}}
    \end{subfigure}

    \begin{subfigure}[t]{0.18\textwidth}
    \centering
    \includegraphics[height=0.085\textheight]{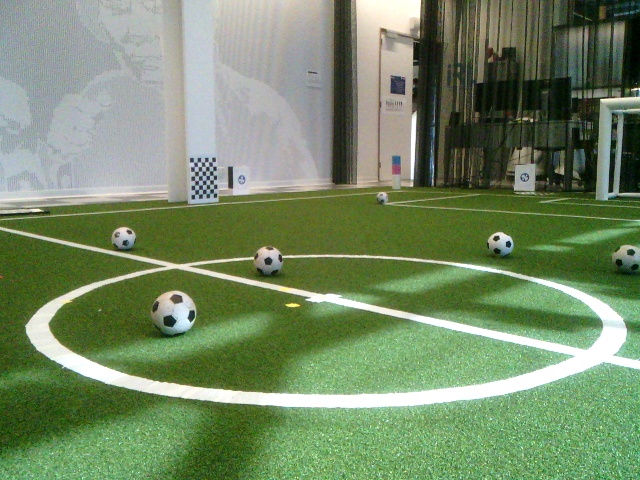}
    \caption{ \label{fig:black1}}
    \end{subfigure}
    \vspace*{1em}
    \begin{subfigure}[t]{0.18\textwidth}
    \centering
    \includegraphics[height=0.085\textheight]{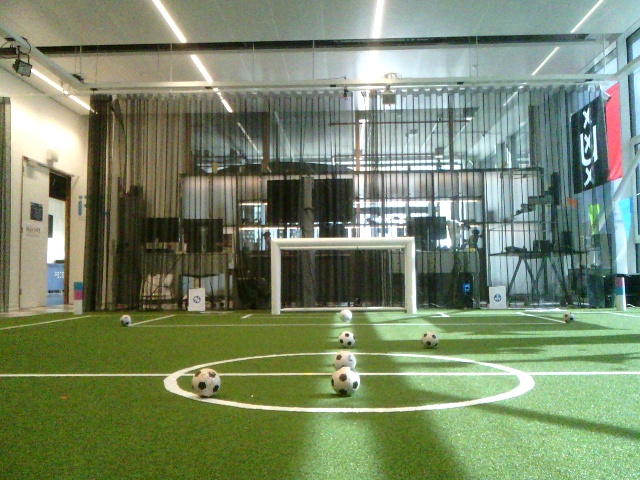}
    \caption{ \label{fig:black2}}
    \end{subfigure}
     \begin{subfigure}[t]{0.18\textwidth}
     \centering
    \includegraphics[height=0.085\textheight]{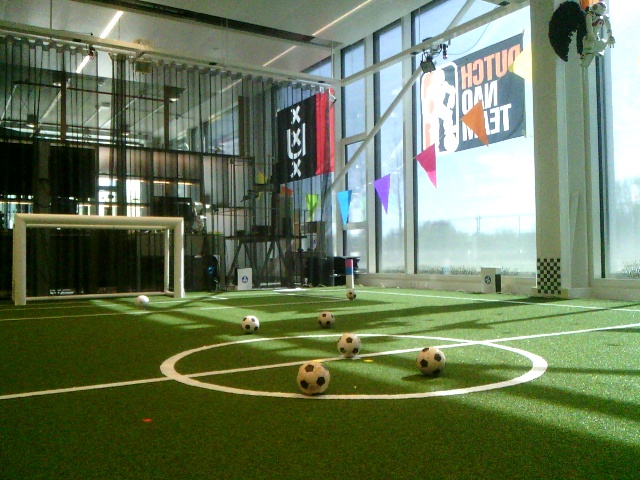}
    \caption{ \label{fig:black3}}
    \end{subfigure}
    \begin{subfigure}[t]{0.18\textwidth}
     \centering
    \includegraphics[height=0.085\textheight]{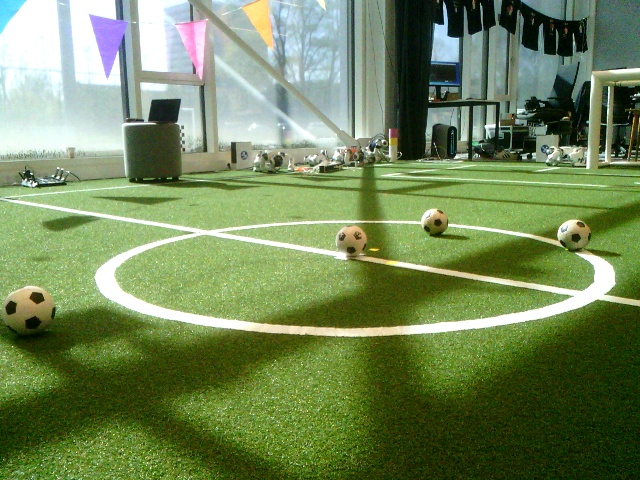}
    \caption{ \label{fig:black4}}
    \end{subfigure}
    \begin{subfigure}[t]{0.18\textwidth}
     \centering
    \includegraphics[height=0.085\textheight]{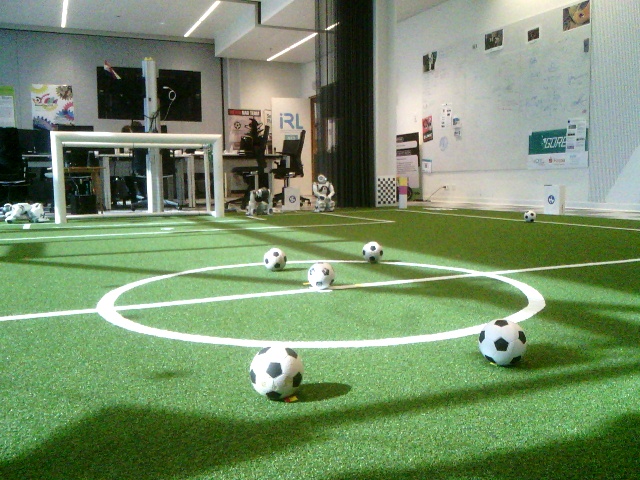}
    \caption{ \label{fig:black5}}
    \end{subfigure}
    \vspace*{-1em}
    \caption{Viewpoints from the 2004 and 2005 SLAM challenge locations}
    \label{fig:viewpoints}
\end{figure}

\begin{figure}[!b]{}
     \centering
  %
    \begin{subfigure}[t]{0.7\textwidth}
    \centering
    \hspace*{-4.0em}
    \includegraphics[height=0.275\textheight]{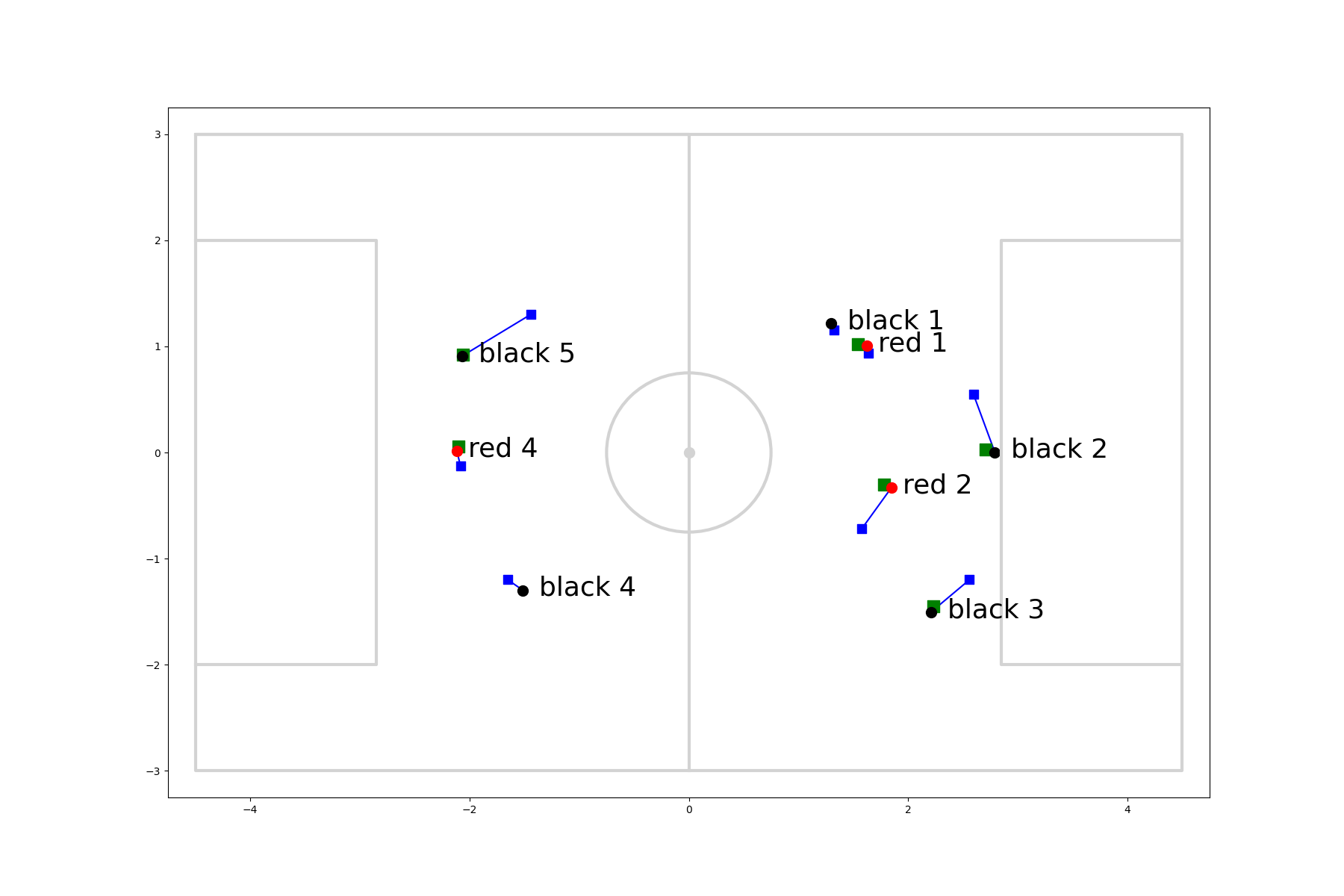}

    \caption{ \label{fig:top-view}}
    \end{subfigure}
    \begin{subfigure}[t]{0.135\textwidth}
    \centering
      \hspace*{-2.0em}
    \includegraphics[height=0.165\textheight]{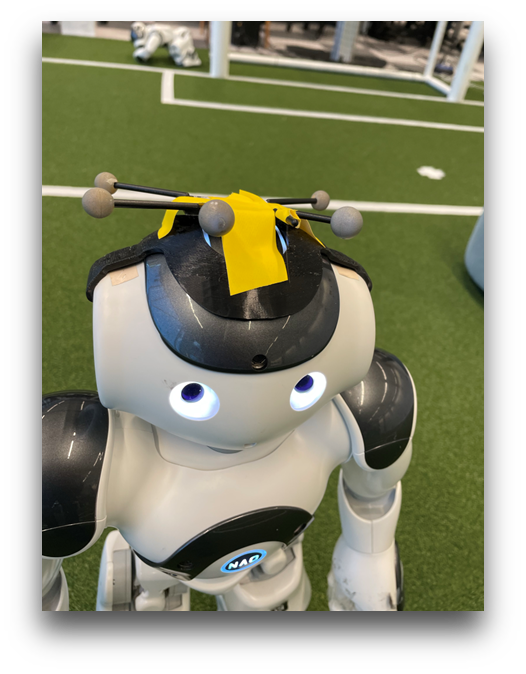}
    \caption{ \label{fig:optitrack-cap}}
    \end{subfigure}
    \caption{Top view of the 2004 and 2005 SLAM challenge locations and a Nao robot with an OptiTrack rigid body marker}
\end{figure}

The soccer fields of 2004 and 2005 were smaller than the current field, so all points are sampled near the center (see Fig.~\ref{fig:top-view})
The collected data is publicly available\footnote{\url{https://github.com/physar/landmark_based_slam_dataset}}. As can be seen from Fig.~\ref{fig:viewpoints}.(f-j), the black points from the 2005 SLAM challenge are slightly further from the center, so here the center circle is always fully visible. Also the weather conditions  changed; the sun was casting shadows when the 2005 SLAM locations were recorded.  

To provide ground truth, the 
Nao robot was equipped with a rigid body marker of the OptiTrack system (see Fig.~\ref{fig:optitrack-cap}), consisting of eight Flex 13 cameras, which can record movements with a 3D accuracy of $\pm 0.20$ mm in a tracking area of $\pm~ 8m\times4m$ \cite{Furtado2019}.

The estimates of location of the observations made on the 10 marked positions from the 2004 and 2005 SLAM 
challenge are summarized in Table~\ref{tab:observations2004} and \ref{tab:observations2005}. As can be seen from Table~\ref{tab:observations2004}, no predictions were made for red point 3 and point 5, because the center circle was not in full view. Also can be seen from the two tables combined that in half of the cases (4 of the 8 marked positions) a reasonable good estimate is given, while for the other half the estimate is clearly off (indicated with a red font). Yet, this clearly wrong location-estimates can be filtered out by setting a threshold on the predicted height, for instance by the constraint that the height as to be in the range $(+0.440,+0.550)$.

\vspace*{-3em}
\begin{table}[!htb]
\caption{Predictions of the observations at the 2004  SLAM challenge locations}
\begin{tabular}{ m{5.5cm} m{1.6cm} m{1.6cm} m{1.6cm} m{0.6cm} }
 & $x$ & $y$ & $h$ & \\
 \textbf{2004 SLAM challenge} \\
 \textbf{red point 1} & $+1.600$ & $+1.000$ &   & m \\
 OptiTrack reference & $+1.626$ & $+1.004$ & $+0.449$ & m \\
prediction with $g$-vector from IMU & $+1.637$ & $+0.939$ & $+0.483$ & m \\
prediction with visual $g$-vector & $+1.545$ & $+1.023$ & \textcolor{red}{+0.369} & m \\
\\
 \textbf{red point 2} & $+1.800$ & $-0.300$ &  & m \\
OptiTrack reference & $+1.849$ & $-0.330$ & $+0.452$ & m \\
prediction with $g$-vector from IMU & \textcolor{red}{$+1.576$} & \textcolor{red}{$-0.721$} & \textcolor{red}{$+0.364$} & m \\
prediction with visual $g$-vector & $+1.784$ & $-0.306$ & $+0.468$ & m \\
\\
 \textbf{red point 3} & $+0.500$ & $-1.000$ &  & m\\
\\
 \textbf{red point 4} & $-2.100$ & $\;\;\; 0.000$ &   & m \\
OptiTrack reference & $-2.115$ & $+0.014$ & $+0.452$ & m \\
prediction with $g$-vector from IMU & $-2.083$ & $-0.125$ & $+0.448$ & m \\
prediction with visual $g$-vector & $-2.100$ & $+0.056$ & \textcolor{red}{$+0.390$} & m \\
\\
 \textbf{red point 5} & $-1.000$ & $+0.500$ &  & m\\
\end{tabular}
\label{tab:observations2004}
\end{table}

Yet, a reliable observation prediction in only half of the cases is not good enough to be useful as input for a localization algorithm. To test whether the origin of this failure comes from the IMU measurement of the $g$-vector, we also estimated the $g$-vector from the goal-posts, when they were visible in the field of view. 
This improved the prediction of the observation locations considerably. With a visual $g$-vector we found the following errors between the prediction and the reference measurement from the OptiTrack system
as indicated in Table~\ref{tab:visual}.

\vspace*{-2em}
\begin{table}[!hbt]
\caption{Difference between predictions with visual $g$-vector and \\ the OptiTrack reference system}
\begin{tabular}{ m{5.5cm} m{1.6cm} m{1.6cm} m{1.6cm} m{0.6cm} }
& $x$ & $y$ & $h$ & \\
\textbf{mean absolute error} 
& 0.044  & 0.030 & 0.076  & m\\
\textbf{standard deviation} 
& 0.037 & 0.017  & 0.069 & m
\end{tabular}
\label{tab:visual}
\end{table}
\vspace*{-1.5em}

Note that for black point 1 and point 3 of the 2005 SLAM challenge (Table~\ref{tab:observations2005}) no goal-posts were in the field-of-view, so visual $g$-vector could be estimated. Also note for red point 1 and point 4 of the 2004 SLAM challenge (Table~\ref{tab:observations2004} the $(x,y)$ prediction with the visual $g$-vector is quite good, but that the constraint on the height-estimate would indicate that this observation could better be skipped.

\begin{table}[htb]
\caption{Predictions of the observations at the 2005 SLAM challenge locations}
\begin{tabular}{ m{5.5cm} m{1.6cm} m{1.6cm} m{1.6cm} m{0.6cm} }
 & $x$ & $y$ & $h$ & \\
 \textbf{2005 SLAM challenge} \\
 \textbf{black point 1} & $+1.300$ & $+1.200$ & & m \\
 OptiTrack reference & $+1.296$ & $+1.217$ & $+0.460$ & m \\
prediction with $g$-vector from IMU & $+1.296$ & $+1.157$ & $+0.538$ & m \\
\\
 \textbf{black point 2} & $+2.700$ & $\;\;\; 0.000$ & & m \\
OptiTrack reference & $+2.791$ & $+0.003$ & $+0.472$ & m \\
prediction with $g$-vector from IMU & \textcolor{red}{$+2.596$} & \textcolor{red}{$+0.549$} & \textcolor{red}{$+0.606$} & m \\
prediction with visual $g$-vector & $+2.706$ & $+0.027$ & $+0.597$ & m \\
\\
 \textbf{black point 3} & $+2.200$ & $-1.500$ & & m \\
OptiTrack reference & $+2.207$ & $-1.505$ & $+0.474$ & m \\
prediction with $g$-vector from IMU & \textcolor{red}{$+2.560$} & \textcolor{red}{$-1.199$} & \textcolor{red}{$+0.577$} & m \\
prediction with visual $g$-vector & $+2.230$ & $-1.449$ & $+0.502$ & m \\
\\
 \textbf{black point 4} & $-1.600$ & $-1.200$ &  & m \\
OptiTrack reference & $-1.513$ & $-1.298$ & $+0.454$ & m \\
prediction with $g$-vector from IMU & $-1.654$ & $-1.298$ & $+0.499$ & m \\
\\
 \textbf{black point 5} & $-2.100$ & $+0.900$ & & m \\
OptiTrack reference & $-2.070$ & $+0.907$ & $+0.425$ & m \\
prediction with $g$-vector from IMU & \textcolor{red}{$-1.439$} & \textcolor{red}{$+1.302$} & \textcolor{red}{$+0.301$} & m \\
prediction with visual $g$-vector & $-2.058$ & $+0.919$ & $+0.449$ & m \\
\end{tabular}
\label{tab:observations2005}
\end{table}

In general, a small error in the viewing direction can give large errors in position estimates. The underlying origin of this small error in estimating the $g$-vector by the IMU can be drift of the accelerometer itself, or a buildup of an error the kinematic chain of several joints between orientation of the head relative to the torso (where the IMU is located).


\section{Discussion}

Finding the viewpoint from a limited number of 'known' points depends strongly on the accuracy of the 'known' part. The Inertial Measurement Unit (IMU) should in principal give an accurate estimate of the $g$-vector. Yet, this $g$-vector is measured inside the torso of the robot, which requires a coordinate transformation from the torso to the camera in the head. Kneip \cite{kneip2011bmvc} assumes that this transformation is fixed and accurately established with a calibration procedure \cite{Lobo2007}, which is unfortunately not possible on a moving robot. When the $g$-vector was estimated from the perspective lines in the image, instead of the IMU measurement, the results actually improved. This indicates that our measurement strongly depends on a good estimate of the $g$-vector.

A better estimate could also be accomplished by combining the constraints from multiple known points, as demonstrated by several solutions available in the OpenGV library \cite{Kneip2014}. In this paper only the X-intersections of middle line and circle were used. Yet, the Standard Platform field has many other useful points, such as the L-corners of the penalty area or the T-intersections at the edge of the field. All these type of intersections can be recognized with the YOLO v8 detector. Yet, the field of the view of the Nao robot is limited, so even in most cases not more than two known points are visible. Including the goal-posts as 'known' points would increase the number of candidates, but it remains a challenge to have at least three or four of those points visible. When playing a soccer match with the occlusion of the opponents nearby, the situation will be even worse. Having an algorithm that can give a reasonable estimate on just two 'known' points remains valuable. 

\section{Conclusion}

In this paper the claim that the use of rational trigonometry can reduce the time to calculate a triangulation is reproduced in a robotics application. 
This is  beneficial for a robotics platform like the Nao robot, where computation time is scarce. 

In addition, this paper demonstrates that for a known orientation relative to the $g$-vector, observing two 'known' points in the environment is enough to get an estimate on the pose of the camera. The accuracy depends on 
an accurate measurement of the $g$-vector by either an Inertial Measurement Unit or the vanishing point from two vertical lines. Estimates that that are completely off can be filtered out by setting a boundary of the height estimate; the height of the robot is known, measurements above that height are only possible when the robot is lifted from the ground. 

The estimates of the relative position to two 'known' points can be the input of a full-scale localization or a landmark based simultaneous localization and mapping algorithm, which keeps track of those observations over time, combined a model of the movement of the robot.

\subsection*{Acknowledgement}

 We like to thank Madelon Bernardy for her support in collecting a dataset, including the reference measurements with the OptiTrack system. We like to thank Gijs de Jong for extending the YOLO v8 XL detector with annotated training data for the X-
 T- and L-intersections.

\bibliographystyle{splncs04}
\bibliography{references}

\end{document}